\pdfoutput=1

\documentclass[11pt]{article}
\usepackage{emnlp2022}
\usepackage{times}
\usepackage{latexsym}

\usepackage[T1]{fontenc}

\usepackage[utf8]{inputenc}

\usepackage{microtype}
\usepackage{latexsym}
\usepackage{multirow}
\usepackage{multicol}
\usepackage{graphicx}
\usepackage{booktabs}
\usepackage{subfigure}
\usepackage{caption}
\usepackage{url}
\usepackage{amsmath}
\usepackage{amssymb}
\usepackage{amsfonts}
\usepackage{comment}
\usepackage{array}
\usepackage{multirow}
\usepackage{microtype}

\usepackage{enumitem}

\usepackage{xcolor}

\title{Varifocal Question Generation for Fact-checking} 
\author{Nedjma Ousidhoum$^*$
  Zhangdie Yuan$^*$  
 Andreas Vlachos\\
 Department of Computer Science and Technology \\
 University of Cambridge \\
 \texttt{ndo24,zy317,av308@cam.ac.uk} \\
}
 
\date{}

\begin{document}

\maketitle
\def\thefootnote{*}\footnotetext{Equal contribution.}
\begin{abstract}
Fact-checking requires retrieving evidence related to a claim under investigation. The task can be formulated as question generation based on a claim, followed by question answering.
However, recent question generation approaches assume that the answer is known and typically contained in a passage given as input,
whereas such passages 
are what is being sought when verifying a claim.
In this paper, we present {\it Varifocal}, a method that generates questions based on different focal points within a given claim, i.e.\ different spans of the claim and its metadata, such as its source and date.
Our method outperforms previous work on a fact-checking question generation dataset on a wide range of automatic evaluation metrics.
These results are corroborated by our manual evaluation, which indicates that our method generates more relevant and informative questions.
We further demonstrate the potential of focal points in generating sets of clarification questions for product descriptions.
\end{abstract}

\section{Introduction}
\begin{figure*}[!t]
    \centering
    \fbox{\includegraphics[scale = .3]{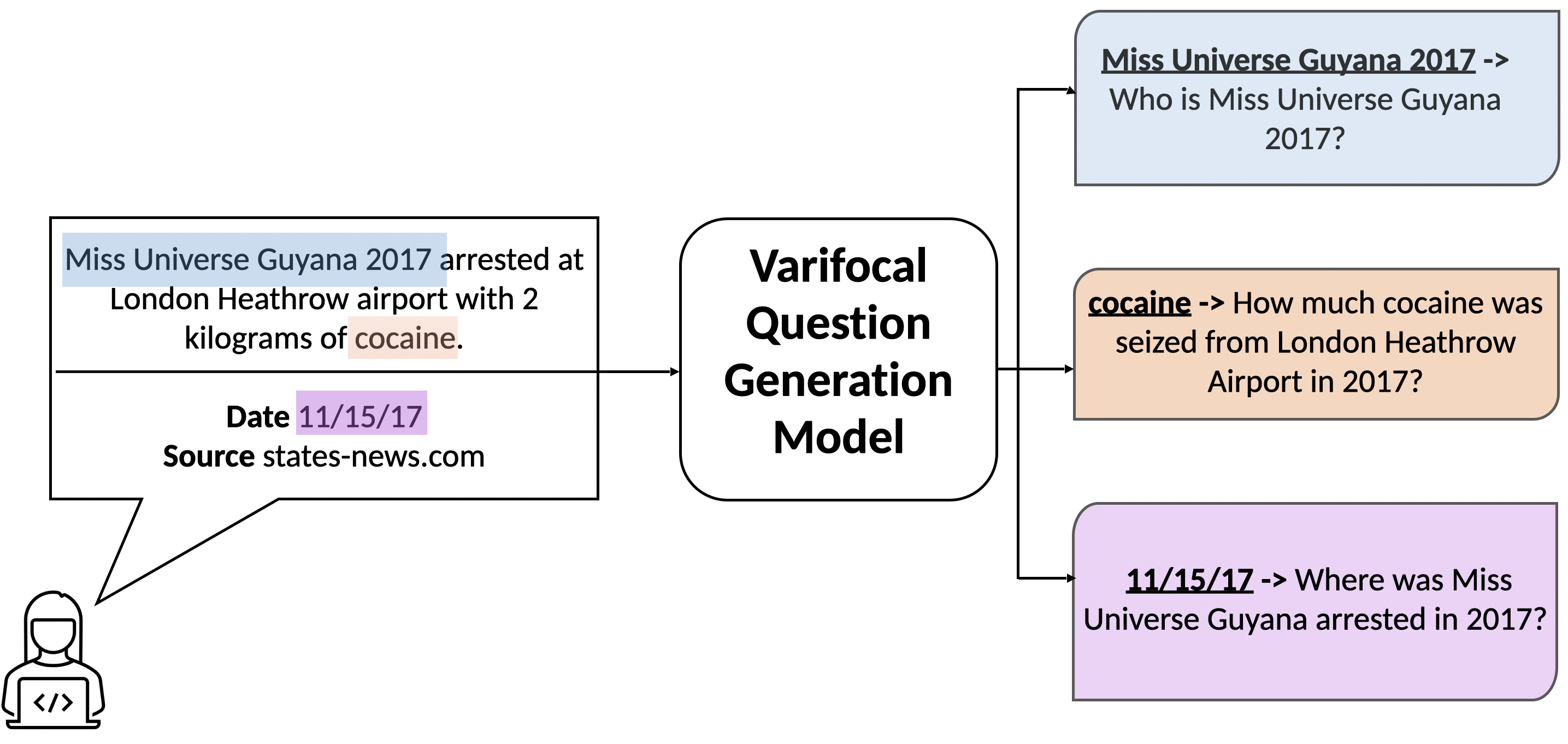}}
    \caption{The architecture of Varifocal. We use a dependency parser to extract the different focal points, i.e.\ spans, then generate questions based on them. We rank the generated questions using a re-ranker and return the top $n$ questions.
The example in the figure was generated by our system. We show three highlighted focal points along with the (output) questions they led to.}
    \label{fig:pipeline}

\end{figure*}

The growing amount of information online and its impact have increased the need for fact-checking, i.e.\ judging whether a claim is true or false. 
 To determine the truthfulness of a claim, fact-checkers need to answer questions related to the claim, world knowledge within its time frame, local politics, etc.\ 
 \cite{graves2017anatomy}. Using questions and answers has also been shown to be an effective way of conveying fact-checks.
For instance, \citet{altay2021covid_clickbot} found that presenting information related to COVID-19 as answers to questions improved attitudes towards vaccination more than merely presenting the relevant facts.

As professional fact-checkers can spend a day to verify a single claim depending on its complexity \cite{adair2017progress,hassan-et-al-2017-check-worthy-claims}, there has been a growing focus on how to accelerate the fact-checking process via automation \cite{cohen2011computational,graves2020discipline}. \citet{fan-etal-2020-fact-checking-briefs} showed that generating questions and answering them reduces the time spent on verification by approximately 20\%.
Fact verification questions tackle information that is missing from the claim, which renders the generation task challenging yet useful for assisting professionals.

Previous work on question generation assumes that the answer is known, typically contained in a passage given in the input
\cite{Rajpurkar-et-al-squad,duan-etal-2017-qg-for-qa,wang2017gated}. 
Such passages, though, are what is being sought when fact-checking a claim.
The only exception is recent work on clarification questions \cite{RaoD19-answer_based_adverserial,majumder_etal_2021_ask_whats_missing_and_useful}. 
However, work in this area examines a specific narrow domain where a limited number of questions can be asked, e.g.\ questions related to product descriptions in the Amazon dataset of \citet{mcauley2016addressing}, or dialogues in the Ubuntu dataset \cite{lowe2015ubuntu}, which is not the case in fact-checking. Fact-checking questions are more diverse and may rely on the experience and intuition of the fact-checker as they aim to scrutinize every piece of information within the claim. They can be as generic as questions about general definitions or as specific as those tackling details about a one-time event.

 In this paper, we propose an approach that generates questions for claim verification, which we name {\it Varifocal}. It uses focal points from different spans of the claim as well as its metadata, i.e.\ its source, and its date.
Each focal point guides the generator to question a different part of the claim; e.g.\ in Figure~\ref{fig:pipeline} when \textit{``Miss Universe Guyana 2017''} is used as the focal point, the question generated is about who she is. 
 
 We evaluate our approach on the QABriefs dataset introduced by \citet{fan-etal-2020-fact-checking-briefs} using a wide range of automatic metrics, and show that Varifocal performs the best among the different systems considered.
 In addition, we conduct a human evaluation on questions generated by four different systems and gold standard questions based on four criteria: a)\ intelligibility, b)\ clarity, c)\ relevance, and d)\ informativeness.
 The results show that Varifocal generates more intelligible, clear, relevant, and informative questions than the other systems, corroborating the results of the automatic evaluation.\footnote{Our code is available on \url{https://github.com/nedjmaou/Varifocal_Fact_Checking_QG}}
 
 Finally, we apply Varifocal to generating sets of clarification questions on Amazon product descriptions \cite{mcauley2016addressing}, where it shows competitive performance against methods that generate single questions while having other ones in the set as part of the input \cite{majumder_etal_2021_ask_whats_missing_and_useful}.

\section{Related Work}\label{sec:related_work}
The main piece of previous work on question generation for fact-checking is by \citet{fan-etal-2020-fact-checking-briefs}. They proposed the QABriefs dataset, which consists of claims with manually annotated question-answer pairs containing additional information about the claims (e.g.\ the exact definition of a term, the content of a bill, details about a political statement or a vote). The QABriefs dataset contained questions asked by crowd-workers, who had to read both the claim and its fact-checking article. \citet{fan-etal-2020-fact-checking-briefs} presented the QABriefer model, which generates a set of questions conditioned on the claim, searches the web for evidence and retrieves answers.
However, they evaluated the questions generated only using BLEU scores without conducting a human evaluation. More recently, \citet{yang2021explainable} addressed the problem of explainability in fact-checking through question answering using the Fool Me Twice corpus \cite{eisenschlos2021fool}. They generated questions from the claim, retrieved answers from the evidence, and compared them to the generated ones.
Yet, they did not evaluate their question generation process, and assumed that the evidence is given as input to generate the questions, which is unrealistic since the questions are typically used for evidence retrieval.

Other related work includes \citet{saeidi2018interpretation} who introduced a dataset containing 32k instances of real-world policies, crowd-sourced fictional life scenarios, and dialogues in order to reach a final yes/no answer. The policies were given as input and were explicitly stating what information needed to be asked for, and the questions had to have a yes or no answer. Neither of these hold in fact-checking, where questions are not usually answered by yes or no, and the information to be searched for is not known in advance.
More recently, \citet{majumder_etal_2021_ask_whats_missing_and_useful} presented a method to generate clarification questions.
They built a two-stage framework that identifies missing information using the notions of \textit{global} and {\it local schemas}. The {\it global schema} was built using filtered key phrases extracted from contexts that were part of the same class of the data, e.g.\ a class of similar products in the Amazon data \cite{mcauley2016addressing} and similar dialogues in the Ubuntu dataset \cite{lowe2015ubuntu}, whereas the {\it local schema} was built using one given context, and they defined the missing information as the difference between the global and the local schema. The extraction of comparable schemas across different contexts was possible due to the repetitive nature of the datasets considered, e.g.\ the descriptions of products of the same type such as laptops allow the prediction of potentially missing properties which need clarification, in contrast to fact-checking claims which are less repetitive.

The standard sequence-to-sequence architecture \cite{sutskever2014sequence} is typically used in question generation approaches \cite{du-etal-learning-to-ask-2017,zhou2017neural}. 
Although answer-aware approaches allow for the generation of multiple questions conditioned on the same passage \cite{sun2018answer}, providing the answer during inference is not possible in fact-checking since one would typically ask questions about what is missing from the claim.
Other work includes question generation for question answering \cite{duan-etal-2017-qg-for-qa}, question generation for educational purposes \cite{heilman-smith-2010-good}, and poll question generation from social media posts \cite{lu-etal-2021-engage}. Furthermore, \citet{hosking-riedel-2019-evaluating} evaluated rewards in question generation, showed that they did not correlate with human judgments, and explained why rewards did not help when using reinforcement learning.

Commonly used evaluation metrics such as BLEU \cite{papineni2002bleu} and ROUGE \cite{lin2004rouge} fall short at correlating with human judgments when evaluating the quality of automatically generated questions~\cite{liu-etal-2016-not-evaluate,sultan-etal-2020-importance,nema2018towards_metric_qg}. \citet{majumder_etal_2021_ask_whats_missing_and_useful} carried a human evaluation based on fluency, relevance, whether the question dealt with missing information, and usefulness. In addition, \citet{cheng-etal-2021-guiding} proposed to assess the quality of automatically generated questions based on whether they were well-formed, concise, answerable, and answer-matching. Similarly, we conduct a human evaluation of the generated questions adapted to fact-checking.

\section{Varifocal Question Generation}
\label{sec:varifocal}

In this section, we describe \textit{Varifocal}, an approach that generates multiple questions per claim based on its different aspects, which correspond to textual spans that we call \textit{focal points}.

Varifocal consists of three components: (1)\ a focal point extractor, (2)\ a question generator that generates a question for each focal point, and (3)\ a re-ranker that ranks the generated questions, removes duplicates and promotes questions that are more likely to match the gold standard ones.

\subsection{Focal Point Extraction}

We consider two types of focal points: contiguous spans from the claim and metadata elements.
For the former, we consider all the subtrees of its syntactic parse tree, thus obtaining more coherent phrases than if we extracted randomly selected n-grams.
In addition, the metadata, which includes (1)\ the source of the claim or the name of the speaker, and (2)\ the date when the claim was made, can be useful in question generation for fact-checking. 
As shown in Figure~\ref{fig:pipeline}, having access to the date of the claim helped the model generate a precise question, i.e.\ Where was Miss Universe Guyana arrested \textit{in 2017}?. 
As the metadata is not part of the claim, we incorporate it using a template. For instance, we combined the claim and metadata of the example shown in Figure~\ref{fig:pipeline} as follows: {\it state-news.com reported on 11/15/17 that Miss Universe Guyana 2017 was arrested at London Heathrow airport with 2 kilograms of cocaine}.

\subsection{Question Generation}

This component takes a claim and its focal points as input and generates a set of questions. Given a claim $c$, the set of all focal points is denoted as $F$, where each focal point $f_i \in F$ is a span in the claim $c$ and its metadata, such as $f_i= [w_{s},...,w_{e}]$ where $s$ and $e$ mark the start and the end of the span, respectively. Then, for each focal point $f_i$, the model generates autoregressively a question $\hat{q_i}$ of $n$ words, as follows:
\begin{equation} \label{Architecture}
    p(\hat{q_i} | c, f_i) = \prod^{n}_{k=1}p
    (\hat{q}_{i_{[k]}} |\hat{q}_{i_{[0:k-1]}}, [\Tilde{c};\Tilde{f_i}])
\end{equation}
$[\Tilde{c};\Tilde{f_i}]$ is the transformer-based encoding of $c$ concatenated to $f_i$. The question generation component in Varifocal is similar to the answer-aware sequence-to-sequence model \cite{sun2018answer}. But, the generator is trained to use focal points instead of answers to the questions that need to be generated. While focal points act similarly to prompts \cite{radford2018improving, brown2020language,
liu2021prompting_survey}, we typically have multiple focal points that are different for each claim, depending on the complexity of its parse tree, as opposed to a small number of fixed prompts, such as one per label in classification tasks.

\subsection{Re-ranking}
\label{ssec:reranker}
After all question candidates are generated (one per focal point), the re-ranker removes duplicated questions in addition to almost identical ones by setting a BLEU score threshold to 0.8.

The re-ranker then scores the remaining questions using a regression model, which assigns a real number score to each candidate.
The more similar the candidate is expected to be to one of the gold questions, the higher score it should receive.

\subsection{Training}
\label{ssec:training}
To train the question generation model, we need focal points paired with the questions that they led to generate. However, most question generation datasets have questions paired with answers instead of focal points. Therefore, we use cosine similarity to match the extracted focal points $f_i \in F$ with the gold answers during training. I.e. \ we calculate $sim(emb(f_i),emb(a_j))$: the cosine similarity between the embeddings of $f_i$ and the gold answer $a_j$.
Following this, we greedily match each answer (and associated question) with the highest scoring focal point. We then remove the latter from the set of focal points available for matching.
In our experiments, we generate the embeddings $emb(f_i)$ and $emb(a_j)$ using Sentence-BERT \cite{reimers2019sentence}.

We train the re-ranker on a holdout split, i.e.\ a small portion of the data that we did not use to build the generator.
Given a claim $c$, the re-ranker $g$ is trained to predict the similarity score that a question would have with the best matching question from the gold standard. Therefore, it considers the maximum sentence similarity of each of the generated questions $\hat{q_i}$ and the gold standard ones $q_j \in \cal Q$, with $\cal Q$ the set of gold questions associated to $c$.
The objective function is expressed as follows:
\begin{equation}
\begin{aligned}
L(D,\theta)= \frac{1}{n}\sum^{n}_{i=1}(y_i - {\hat{y}_i})^2
\end{aligned}
\end{equation}

 where $y_i = \max_{q_j \in \cal{Q}} sim(emb(\hat{q_i}), emb(q_j))$, $\hat{y}_i = g(\theta,\hat{q_i})$ is the score predicted by the re-ranker, $D$ refers to the training data, $n = |D|$, and $sim$ calculates the cosine similarity between the sentence embeddings $emb(\hat{q_i})$ and $emb(q_j)$ of $\hat{q_i}$ and $q_j$, respectively. We use Sentence-BERT \cite{reimers2019sentence} to compute $sim$.

\section{Experimental Setup}\label{experiments}
\subsection{Data}
We train our question generation system on the QABriefs dataset, which contains 7,535 claims with 21,168 questions. We change the splits used by \citet{fan-etal-2020-fact-checking-briefs} to ensure that all the claims in the test set contain metadata. Our training set contains 5,228 claims associated with 14,371 questions, the validation set has 653 claims associated with 1,958 questions, and the test set is composed of 653 claims associated with 1,952 questions. We also reserve a further holdout split of 999 claims for training the re-ranker. 
In our experiments, we use the SpaCy dependency parser \cite{spacy2} to parse the claims and extract the focal points.

\subsection{Models}
In our experiments, we train the following models.
\begin{itemize}[noitemsep,nolistsep]
\item \textbf{BART} For each claim, we use BART \cite{lewis2019bart} to generate a set of questions separated by a separation token {\it [SEP]}. This is a replication of the QABriefer model reported by \citet{fan-etal-2020-fact-checking-briefs}. 
\item \textbf{SQuAD} We train an answer-aware question generation model \cite{sun2018answer} on the SQuAD dataset \cite{rajpurkar2018know}
without further fine-tuning on the QABriefs dataset. However, we use focal points instead of the answers. We extract the focal points using the method described in Section~\ref{ssec:training}.
\item \textbf{Varifocal} We pretrain the model on SQuAD and fine-tune it on the QABriefs dataset. The models take tuples of the form $(c, f_i, q_i)$ as input.
\item \textbf{Varifocal+Meta} We pretrain the model on SQuAD and fine-tune it on the QABriefs dataset with the metadata associated with each claim.
\end{itemize}
We generate different datasets for training a re-ranker for each of the models considered, i.e.\ SQuAD, Varifocal, and Varifocal+Meta, following Section~\ref{ssec:training}.
\subsection{Automatic Evaluation}
 \begin{table*}[!t]
\small
    \centering
    \begin{tabular}{ccccccccc}
       \toprule
        {\bf System} & {\bf BLEU-2}&{\bf BLEU-4} &{\bf chrF} & {\bf METEOR}  & {\bf ROUGE-1} & {\bf ROUGE-2} & {\bf ROUGE-L} &{\bf TER} \\
        \midrule
         {\bf BART}& 25.63 &	11.83&	37.25&	31.57&	33.66&	14.02&	33.34&	0.804\\
         {\bf wh-BART}& 21.88 &	10.54&	40.1&	33.62&	29.97&	13.19&	29.46&	0.8697\\
         {\bf SQuAD} &25.85&	11.95&	38.60&	30.67&	32.70&	13.63&	31.84&	0.809\\
         {\bf Varifocal} &29.98&	15.17&	41.12&	34.84&	37.27&	17.64&	36.77&	\textbf{0.755}\\
         {\bf Varifocal+Meta} & \textbf{30.18}&	\textbf{15.54}&	\textbf{43.17}&	\textbf{37.02}&	\textbf{38.19}&	\textbf{18.37}&	\textbf{37.59}&	0.764\\ 
         \bottomrule
    \end{tabular}
    
    \caption{Automatic evaluation results on the QABriefs test set. For all scores higher is better except for TER.}
    \label{tab:scores}
\end{table*}

We evaluate our question generation models using: 
\begin{itemize}[noitemsep,nolistsep]
    \item \textbf{BLEU} \cite{papineni2002bleu} which evaluates n-gram precision (used $n=2$ and $n=4$),
    \item \textbf{chrF} character n-gram F-score  \cite{popovic-2015-chrf}, 
    \item \textbf{METEOR} \cite{banerjee2005meteor} which is based on unigram precision and recall as well as stemming and synonymy matching for similarity,
    \item \textbf{ROUGE} \cite{lin2004rouge} which evaluates n-gram overlap (used $n=1$ $n=2$), 
    \item \textbf{ROUGE-L} which uses the Longest Common Subsequence statistic,
    \item \textbf{TER} \cite {snover2006_ter} which is an error rate based on edit distance.
\end{itemize}

\subsection{Human Evaluation}
\begin{table}[!t]
	\small
    \centering
    \begin{tabular}{lcccc}
    \toprule
         {\bf Avg}&\textbf{I} &\textbf{C} & \textbf{R} &\textbf{Info} \\
         \midrule 
         \textbf{Gold}&\textbf{0.97}&0.91&0.79&1.72\\
        \textbf{SQuAD}&0.83&0.84&0.77&1.91\\
        \textbf{BART}&0.85&0.76&0.67&1.49\\
        \textbf{Varifocal}&\textbf{0.97}&\textbf{0.94}&\textbf{0.93}&\textbf{2.33}\\
        \textbf{Varifocal+Meta}&0.93&0.91&0.89&2.10\\
         \bottomrule
    \end{tabular}
    
    
    \caption{Average of intelligibility (I), clarity (C), relevance (R), and informativeness scores per system based on our human evaluation.}
    \label{tab:question_quality}
\end{table}
We conduct a human evaluation of the generated questions based on a)\ intelligibility, b)\ clarity, c)\ relevance, and d)\ informativeness.
The raters assign a 0/1 score to the intelligibility, clarity and relevance of the question, and a score from 0 to 3 to its informativeness.
We define these criteria in the context of fact-checking as follows. 
\paragraph{Intelligibility}
The question should be fluent, and as long as it is understandable, it does not have to be perfectly grammatical. The intelligibility of a question should be judged without looking at the claim.
This criterion is similar to the {\it fluency} criterion presented by \citet{majumder_etal_2021_ask_whats_missing_and_useful} and to the \textit{good form} criterion proposed by \citet{cheng-etal-2021-guiding}.

\paragraph{Clarity}
Questions should be clear enough to be answered confidently using a search engine. Hence, a clear question should not be too broad and should include some necessary details, such as the date and the name of the speaker, etc. The question remains clear if these details can be induced by looking at the claim.

\paragraph{Relevance}
A generated question is only relevant if it mentions entities related to the claim. The entities can either be mentioned in the claim or the metadata since we use the latter to train a question generation system.

\begin{table*}[!t]
    \small
    \centering
    \begin{tabular}{cc|c|c|c|c}
    \toprule
    \textbf{Claim 1}& We have trade deficits with almost every country. &\multicolumn{4}{c}{\textbf{Scores}}\\ \cline{3-6}
    &\textit{(Donald Trump, 7/28/17)}&\textbf{I}&\textbf{C}&\textbf{R}&\textbf{Info}\\ \midrule
     \color{violet}\textbf{BART}&  \color{violet}What was the trade deficit with China as Secretary of State?&0&0&0&0\\ \midrule
    \color{magenta}\textbf{Varifocal} & \color{magenta}\textit{trade} $\longrightarrow$What is the trade deficit with China?&1&1&1&1.67\\ \midrule
    \color{violet}\textbf{BART}&  \color{violet}What are the trade deficits with almost every country?&0.33&0&0&0\\ \midrule
     \textbf{Claim 2} & Black home ownership just hit the highest level it has ever been in the &\multicolumn{4}{c}{\textbf{Scores}}\\ \cline{3-6}
     &history of our country. \textit{(Donald Trump, 12/8/17)} &\textbf{I}&\textbf{C}&\textbf{R}&\textbf{Info}\\ \midrule
      \color{magenta}\textbf{Meta+Varifocal}& \color{magenta}\textit{of our country}$\longrightarrow$ \\
      &\color{magenta}What is the highest black home ownership in the history of the US?&1&1&1&3\\ \midrule
      \color{cyan}\textbf{Gold}&\color{cyan}How true is it that black ownership is at its highest level ever?&1&1&1&1.33 \\ \midrule
      \color{violet}\textbf{BART}&  \color{violet}What is the average black home price in the U.S.?&0.33&0.33&0.33&1\\
      \midrule
     \textbf{Claim 3}& \#Broward elections department has a history of violating the law.&\multicolumn{4}{c}{\textbf{Scores}}\\ \cline{3-6}
     &\textit{(Marco Rubio, 11/8/18)}&\textbf{I}&\textbf{C}&\textbf{R}&\textbf{Info}\\
\midrule
 \color{magenta}\textbf{Meta+Varifocal}& \color{magenta} \textit{\#Broward elections department} $\longrightarrow$ \\
 &\color{magenta}Did the Broward elections department violate the law?&1&1&1&3\\ \midrule
  \color{cyan}\textbf{Gold}& \color{cyan}How will Rubio oversee that the election is done correctly?&1&1&0.67&0.67\\ \midrule
 \color{magenta}\textbf{Meta+Varifocal}& \color{magenta}\textit{1/8/18} $\longrightarrow$ \\
 & \color{magenta}When did the Broward elections department violate the law?&1&1&1&2\\ \midrule
 \color{violet}\textbf{BART}& \color{violet}What is the history of violating the law?&1&1&0&0\\
\bottomrule

    \end{tabular}
    \caption{Examples of annotated questions that are either generated by one of the different systems or present in the gold set of questions. We show examples with different averaged human ratings for 1)\ intelligibility (I), 2)\ clarity (C), 3)\ relevance (R) and 4)\ informativeness (Info) scores. The focal points associated with the questions generated by the Varifocal systems are also presented (in the text preceding the arrow).}
    \label{tab:generated_questions}
\end{table*}

\paragraph{Informativeness} 
An informative question should return answers providing information that helps us judge the veracity of the claim. The informativeness of a fact-checking question depends on the nature of the claim. For instance, if the claim is a quote, a question that focuses on the person or entity who made the statement can be informative. On the other hand, if the claim is about an event, then an informative fact-checking question may be about the event itself. 
While questions should not directly ask whether the claim is true or false, a yes/no question which is necessary to reach a verdict is informative.
We use a 4-point Likert scale to assess the informativeness of a question. A question can be: 
\begin{enumerate}[noitemsep,nolistsep]
    \item \textbf{uninformative} i.e.\ useless \textit{(score = 0)},
    \item \textbf{weakly informative} i.e.\ unlikely to be helpful but we do not mind having it generated by the system \textit{(score = 1)}, 
    \item \textbf{potentially informative} or somewhat useful, i.e.\ a question that could be helpful depending on the context of the claim. For instance, a question whose answer is in the claim can be informative if it is worth verifying \textit{(score = 2)}, 
    \item \textbf{informative} i.e.\ a question to which the answer is crucial for the fact-check \textit{(score = 3)}.
\end{enumerate}
An informative question is intelligible, clear, and relevant.
Informativeness differs from the {\it missing information} criterion defined by \citet{majumder_etal_2021_ask_whats_missing_and_useful} in that it is not defined against a predefined schema such as a product description.

We asked three volunteers to assess the quality of the questions. The raters are researchers in NLP (not involved in the paper) working in an English-speaking institution. They were assigned ten questions per claim, i.e.\ two gold questions associated with the claim in the QABriefs dataset and two questions generated by each of our models: SQuAD, BART, Varifocal and Varifocal+Meta. We hid the name of the systems which generated the questions and their ranking from the raters.

\section{Results}

\subsection{Automatic Evaluation}\label{sec:automatic_eval}

The results presented in Table~\ref{tab:scores} show consistency with respect to which system performs the best despite the variation in their rankings. 
The Varifocal systems outperform BART and SQuAD by around 4 ROUGE-1 points, >4 METEOR points, and show lower error rates based on TER. 
Interestingly, SQuAD slightly outperforms BART despite not being trained on the QABriefs dataset. However, SQuAD uses focal points to generate questions, which indicates their potential usefulness.
We notice a minor difference between Varifocal and Varifocal+Meta except for TER, where Varifocal (without metadata) performs slightly better.
To further assess the potential of focal points, we experimented with the BART question generator . We forced a BART model (wh-BART) to initiate the generated questions with the most common question words in the QABriefs dataset, typically wh-words such as \textit{What}, \textit{Why}, \textit{How}, etc. (see Figure~4 in \citet{fan-etal-2020-fact-checking-briefs}). While this resulted in scores comparable to those of BART and SQuAD, these were always lower than the scores achieved by Varifocal, further demonstrating that focal points provide guidance beyond the question type.

\subsection{Human Evaluation }\label{human_eval_res}

The raters evaluated a total of 250 questions generated for 25 different claims. 
As they assigned Boolean values to intelligibility (I), clarity (C), and relevance (R), similar distributions with minor disagreements led to low Fleiss-$\kappa$ and Krippendorff-$\alpha$ scores.
The Fleiss-$\kappa$ scores are 0.48 for intelligibility, 0.43 for clarity, 0.32 for relevance and  0.26 for informativeness. However, as reported in previous work, chance-adjusted scores can be low despite a high agreement due to their inappropriateness when assessing variables with imbalanced marginal distributions \cite{randolph2005free,falotico2015fleiss,yannakoudakis2015evaluating,matheson2019we}. Therefore, we have computed the free marginal multi-rater kappa scores \cite{randolph2005free}, which are equal to 0.88 for intelligibility, 0.74 for clarity, 0.58 for relevance and 0.32 for informativeness. 

Unsurprisingly, intelligibility is the least subjective criterion and is therefore the one on which the raters disagree the less, followed by clarity and relevance. On the other hand, we report a relatively low agreement on the informativeness criterion.
Table \ref{tab:generated_questions} shows examples of questions generated by the different models and how they were rated.

The results of our human evaluation corroborate the scores of the automatic metrics presented in Section \ref{sec:automatic_eval}. Overall, Varifocal and Varifocal+Meta generate the best questions on average based on the four criteria. 
Moreover, they seem to generate more relevant and informative questions even when compared to the gold ones.
In fact, some gold questions consider non-trivial prior knowledge about a claim and, thus, are sometimes annotated as irrelevant, i.e.\ they mention entities and events that are not part of the claim and the metadata. This is due to the fact that when building the QABriefs dataset, the annotators had to read the fact-checking article associated with the claim. As a result, the annotators have sometimes assumed the veracity/falsehood of a claim when asking questions and mentioned entities and events that only appeared in the fact-checking article.
For instance, the gold question of claim 3 shown in Table~\ref{tab:generated_questions} refers to an act (overseeing elections) not mentioned in the claim. 
\begin{figure}
    \centering
    \fbox{\includegraphics[scale = .27]{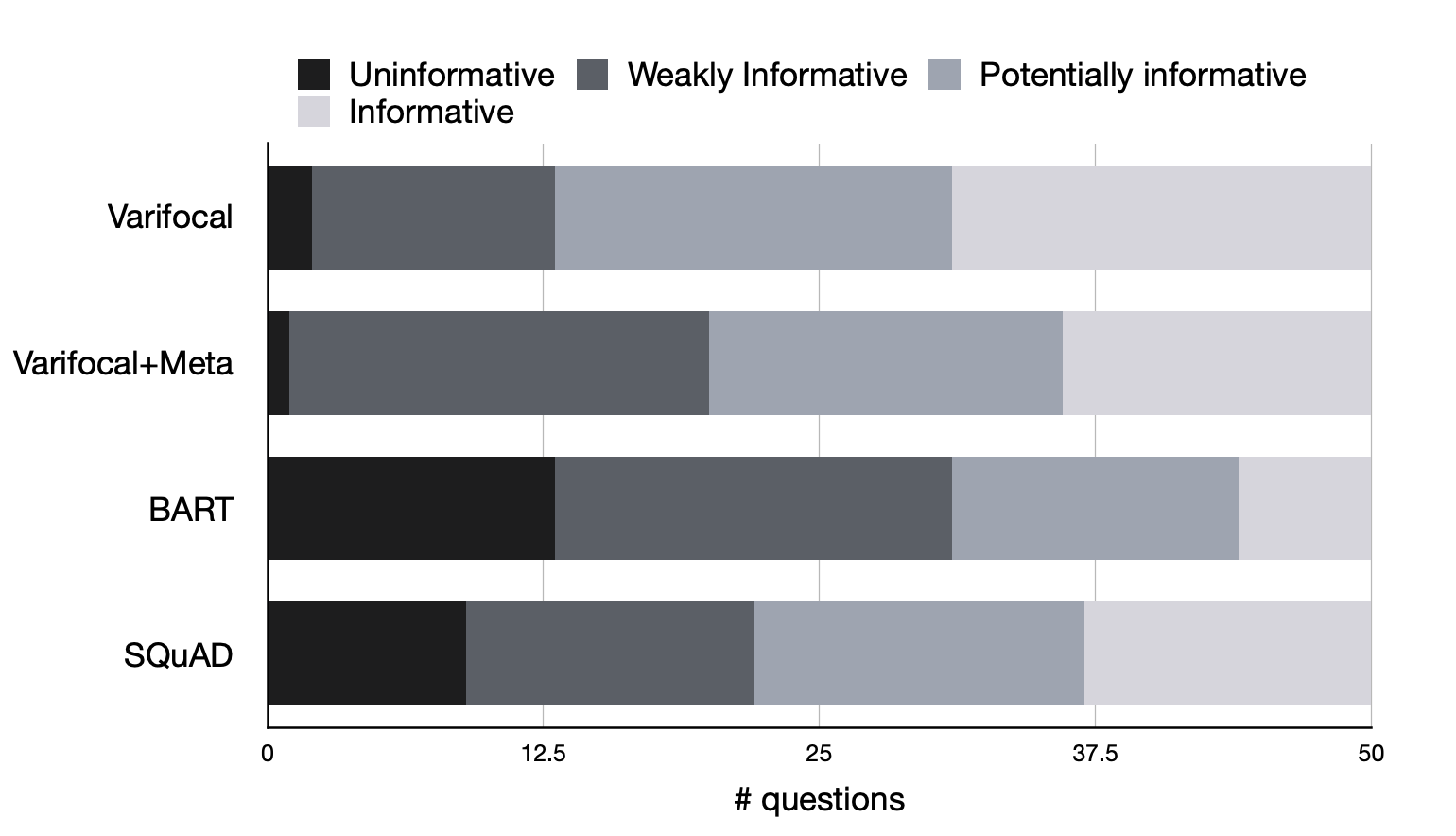}}
    \caption{The distribution of informativeness scores across the different systems. Brighter means better.}
    \label{fig:dist_info}
\end{figure}
We observe the most considerable difference in the high informativeness range (score=3), as shown in Figure \ref{fig:dist_info}. The BART model, i.e.\ the approach used in the QABriefer, fails noticeably to generate fact-checking questions of high informativeness. A relatively large percentage of BART questions were found to be ineffective for fact-checking since an informative question needs to be intelligible, clear, and relevant. Hence, the unclear questions were uninformative by default per our guidelines.
\begin{table*}[!t]
\small
    \centering
    \begin{tabular}{cc}
    \toprule
        \textbf{Claim} &Says Donald Trump promised ``the mass deportation of \underline{Latino families}.'' \\
        \textbf{Question}& \color{magenta}Did Trump promise mass deportation of Latino families?  $\longrightarrow tag(f_i)=pobj$ \\ \midrule
        \textbf{Claim} & Says \underline{NRA head Wayne LaPierre} said, We believe in absolutely gun-free, zero-tolerance, totally safe schools.\\ & That means no guns in America’s schools. Period.\\
        \textbf{Question}& \color{magenta} What is Wayne LaPierre's stance on guns in schools?  $\longrightarrow tag(f_i)=nsubj$\\ \midrule
        \textbf{Claim} & Bay Area liberals have given \underline{more} to Jon Ossoff's campaign \underline{than people in Georgia}.\\
        \textbf{Question}& \color{magenta} How much money do people in Georgia give to Jon Ossoff? $\longrightarrow tag(f_i)=prep$ \\ \midrule

    \end{tabular}
    \caption{Examples of questions with informativeness scores = 3 generated for focal points with tags $\in \{nsubj, pobj, prep\}$. The focal points are underlined in their respective claims.}
    \label{tab:examples_subsets_crucial}
\end{table*}

\subsection{Evaluation on the Amazon Dataset}
To evaluate the potential of Varifocal and focal points outside fact-checking, we used the Amazon dataset of \citet{mcauley2016addressing}. 
We trained our system (pretrained on SQuAD) on Amazon product descriptions only. Following Section~\ref{ssec:training}, we extracted the focal points using SpaCy \cite{spacy2}, trained a re-ranker, and then generated and ranked multiple questions per product.
We achieved a BLEU-4 score of 13.2 using only the focal points as guidance. In contrast to the reported experimental setup by \citet{majumder_etal_2021_ask_whats_missing_and_useful}, who used previously asked questions as part of the input, effectively predicting the missing question in a set, we generate all questions given the product descriptions only.
Their best model was trained on product descriptions, questions on the product, and previously asked questions about related products required to model missing information based on the notion of schema, i.e.\ \textit{global schema - local schema}. This model achieves a BLEU-4 score of 18.55 and conditions BART on a usefulness classifier during decoding using Plug and Play Language Models (PPLMs) \cite{dathathri2019pplm}. 

Nevertheless, among their baselines, they also trained a Transformer model that achieved a 12.89 BLEU score. This baseline was outperformed by Varifocal, although our system generated each question given the product description without access to the other questions asked about the product.
 
In conclusion, although modelling missing information using a schema can be useful, it can only be applied to a narrow domain, such as similar product descriptions. It is also worth noting that generating fact-checking questions can be harder than generating clarification ones. For instance, for the claim \textit{``\#Broward elections department has a history of violating the law."} shown in Table \ref{tab:generated_questions}, the question \textit{``What is the history of violating the law?"} can be considered a clarification question, but not a good fact-checking one.

\section{Analysis}\label{analysis}
To investigate the quality of the different focal points and how they guide the question generation process (1)\ we extracted focal points that are named entities only and those that are not, then automatically evaluated the performance of Varifocal on each of these subsets; (2)\ we compared the average human rating scores of all focal points to the ones achieved by subsets of focal points referring to specific syntactic tags. We show examples of questions generated for focal points with some of these tags in Table \ref{tab:examples_subsets_crucial}.

\begin{table}[!t]
\small
    \centering
    \begin{tabular}{cccc}
    \toprule
        &\textbf{BLEU-4}&\textbf{ROUGE-2}&\textbf{METEOR}  \\ \midrule
        \textit{NE} &15.44&35.15&17.48\\
        \textit{Non NE}&\textbf{16.13}&\textbf{36.88}&1\textbf{8.33}\\
\bottomrule
    \end{tabular}
    \caption{BLEU-4, ROUGE-2 and METEOR scores achieved for focal points that are named entities (NE) only vs. other focal points (Non NE).}
    \label{tab:focal_points_subsets}
\end{table}

Table \ref{tab:focal_points_subsets} shows that not considering named entities as focal points performs better than only considering named entities. 
This observation is especially relevant for fact-checking since named entities are often used in heuristics that generate questions \cite{lewis2021paq}. On the other hand, this also proves that reducing the number of focal points does not necessarily affect the quality of the question generation. 

When analysing the results, we observed that focal points whose syntactic roles are \textit{nsubj}, \textit{dobj}, \textit{pobj}, \textit{prep} and \textit{compound} seem to lead to the best questions. We, therefore, examined their average informativeness scores according to the human raters. As shown in Table \ref{tab:avg_info_scores}, the top 5 tags have an average informativeness score that is >2.

\section{Limitations and Future Work}
\begin{table}[!t]
\small
    \centering
    \begin{tabular}{cc}
    \toprule
    \textbf{Focal points} &  \textbf{Avg(Informativeness)} \\ \midrule
        \textit{All}&2.33\\
        $tag(f_i)$=nsubj&2.18 \\	$tag(f_i)$=dobj&2.58\\	$tag(f_i)$=pobj&2.49\\
        $tag(f_i)$=prep&\textbf{2.63}\\	
        $tag(f_i)$=compound&2.37\\ \bottomrule
    \end{tabular}
    \caption{Average informativeness scores of the questions generated for focal points $f_i \in F$ whose tags are nsubj, dobj, pobj, prep and compound, respectively. \textit{All} refers to the average informativeness score of all the questions generated by Varifocal that were labeled by our raters.}
   \label{tab:avg_info_scores}
\end{table}

\paragraph{The Complexity of the Claims}

In our experiments on the QABriefs dataset, the maximum number of focal points was 175, as shown in Table \ref{tab:focal_points} despite fact-checking claims being short in length and limited in terms of complexity. As the average number of focal points was 22 (or 28 when using metadata), we over-generated questions causing a fair amount of duplicates that needed to be removed by the re-ranker.
However, as shown in Section \ref{analysis}, using a subset of focal points can be sufficient to achieve a comparable performance, which can help us reduce the running time of our method. 
By knowing in advance the types of focal points that would more likely lead to good questions, we can avoid extracting all the possible ones and reduce the complexity of our method.

\paragraph{Additional Evaluation}

In the future, besides evaluating the quality of individual questions in terms of intelligibility, clarity, relevance and informativeness, we intend to assess the quality of sets of questions. Metrics such as Distinct-2 \cite{li-etal-2016-diversity} used by \citet{majumder_etal_2021_ask_whats_missing_and_useful} assess the lexical diversity of the sets whereas {\it semantic} diversity is the one that is critical for fact-checking \cite{sultan-etal-2020-importance}.  
The evaluation of a set of questions needs to take different dimensions into account and can depend on the nature of the claim for which we generate questions (e.g.\ a statement vs. an opinion).
Furthermore, we plan to assess questions considering the limitations of currently used search engines, i.e.\ whether the question is numerical, comparative, ambiguous, and whether its answer can, therefore, be easily fetched or not.

\begin{table}[!t]
\small
    \centering
    \begin{tabular}{cccc}
    \toprule
        &\textbf{Min}&\textbf{Avg}&\textbf{Max} \\\midrule
         \textbf{Varifocal} &4 & 22 & 169 \\ \midrule
         \textbf{Varifocal+Meta}& 10 & 28 & 175 \\ \bottomrule
    \end{tabular}
    \caption{Maximum, minimum, and average numbers of extracted focal points per system, i.e\ with and without using metadata.}
    \label{tab:focal_points}
\end{table}

\paragraph{Knowledge Beyond the Claim}
One can also argue that only using the claim and the metadata to generate focal points and questions is insufficient for fact-checking. In the future, we propose to predict potential answers to the generated questions using concepts and entities related to different parts of the claim with the help of knowledge bases and search engines.

\section{Conclusion}
We introduced Varifocal, a question generation method for fact-checking that alleviates the absence of full context using focal points. 

Varifocal, with and without metadata, improves the quality of automatically generated questions compared to other systems. It generates intelligible and clear questions, and sometimes questions that can be more relevant and informative than the gold standard ones. 
Moreover, when we used Varifocal to generate sets of clarification questions, it showed comparable results to those achieved by models that generate single questions while having additional ones as part of the input.

In the future, we will consider extensions using additional knowledge related to the different aspects of the claim and assess the overall quality of sets of questions.

\section{Acknowledgements}
We acknowledge Marzieh Saeidi for early contributions to this work. 
We thank Nicolas Patz and colleagues at Deutshe Welle, Andrea Parker, and Bryan Chen for evaluating previous versions of our system. We also thank Michael Schlichtkrull, Pietro Lesci, and Zhijiang Guo for their help and insightful feedback, as well as Afonso Mendes and anonymous reviewers and meta-reviewer for their comments on earlier versions of the manuscript.

Nedjma Ousidhoum is supported by the EU H2020 grant MONITIO (GA 965576). Zhangdie Yuan is supported by the ERC grant AVeriTeC (GA 865958).
Andreas Vlachos is supported by both AVeriTeC and MONITIO.
%


 
\bibliographystyle{acl_natbib}
\bibliography{references}
\newpage
\appendix

\section{Annotation Guidelines}
We are evaluating a question generation system for fact-checking. First, we assess each question independently, then we evaluate the whole set of generated questions.

We assess each generated question based on the following criteria. 

\subsection{Intelligibility}
The question should be fluent but does not have to be perfectly grammatical as long as it is understandable. The intelligibility of a question should be judged without looking at the claim.

\paragraph{Examples of intelligible vs. unintelligible questions}
    \begin{itemize}
    \item \textbf{Unintelligible question} How many 
    less do Florida's teachers pay? (\textit{Incomprehensible.})
    \item \textbf{Intelligible question} What made Rep. Paul Gosar to ask for the arrest of the illegal immigrants? (\textit{Not perfectly grammatical but still intelligible.})
    \item \textbf{Intelligible question} What is the average pay for Florida's teachers?	(\textit{Grammatical and intelligible.})

\end{itemize}

\subsection{Clarity}
Questions should be precise enough to be answered confidently using a search engine regardless of the context. A clear question should not be too broad and should include all the necessary details, such as dates, and names of people/speaker, etc. If the details can be induced by looking at the claim, the question remains clear.
\paragraph{Examples of clear vs. unclear questions}
\begin{itemize}
    \item \textbf{Unclear question} What is the name of the state that New Jersey elects a Republican to the Senate? (\textit{Unintelligible and unclear.})
    \item \textbf{Unclear question} What policies violate federal law? (\textit{Too broad.})
    \item \textbf{Unclear question} What did the author of the bill say about the bill? (\textit{Intelligible but unclear.})

    \item \textbf{Clear question} What is the definition of a sanctuary city?
    \item \textbf{Clear question} What is the United Nations?
    \item \textbf{Clear question} What was the name of the law that separated children from adults entering America?
    \item \textbf{Claim} Apprehension rates at the southern border have plummeted since the 1980s and apprehensions of Mexicans specifically have reached their lowest point in nearly half a century.
    \begin{itemize}
        \item \textbf{Clear question when looking at the claim (otherwise, unclear since the name of the country is not specified)} What was the apprehensions rate at the southern border in the 1980s?
    \end{itemize}

\end{itemize}


\subsection{Relevance}
The generated questions should mention entities that are related to the claim. The entities can either be mentioned in the claim or in the metadata since we may use the latter to train a question generation system. 

\paragraph{Examples of relevant vs. irrelevant questions}
\begin{itemize}
    \item \textbf{Claim} Miss Universe Guyana 2017 arrested at London Heathrow airport with 2 kilograms of cocaine.
    \begin{itemize}
    \item \textbf{Irrelevant} Why would someone make up a fake news story about her hiding cocaine in coffee bags?
    \item \textbf{Irrelevant} Why would someone make up a fake news story about her hiding cocaine in coffee bags?
    \item \textbf{Relevant} Who was the Miss Universe Guyana 2017 arrested at London Heathrow airport with 2 kilograms of cocaine?
    \item \textbf{Relevant} Who is Miss Universe Guyana 2017?
    \item \textbf{Relevant} What is the name of the person arrested at London Heathrow airport?
    \end{itemize}
\end{itemize}

\subsection{Informativeness}
An informative question should return answers that provide information about the claim in order to help us reach a verdict for its veracity. The informativeness of a fact-checking question will depend on the type of the claim. For instance, if the claim is a quote, a question which focuses on the person or entity who made the statement can be an informative one. On the other hand, if the claim focuses on the narration of a certain event, then an informative fact-checking question may focus on the event itself. 
A yes/no question which is useful to reach a verdict is considered to be informative. Questions, however, should not directly ask or imply that the claim is true or false.
Finally, an informative question should not (indirectly) imply that the claim is true or false.

We use a 4-point Likert scale to assess the informativeness of a question. A question can be: 
\begin{enumerate}
    \item \textbf{uninformative} i.e.\ useless (0),
    \item \textbf{weakly informative} i.e.\ unlikely to be helpful but we do not mind to have it generated by the system\textit{(score = 1)} , 
    \item \textbf{potentially informative} or somewhat useful, i.e.\ a question that could be helpful depending on the context of the claim. For instance, a question whose answer is in the claim can be informative if it is worth verifying \textit{(score = 2)}, 
    \item \textbf{informative} i.e.\ crucial \textit{(score = 3)}.

\end{enumerate}

\paragraph{Examples questions scored according to their informativeness}
\begin{itemize}
    \item \textbf{Claim} Miss Universe Guyana 2017 arrested at London Heathrow airport with 2 kilograms of cocaine.
    \begin{itemize}
    \item \textbf{Irrelevant and uninformative } Why would someone make up a fake news story about her hiding cocaine in coffee bags?
    \item \textbf{Relevant and weakly informative} What is the name of the person arrested at London Heathrow airport?
    \item \textbf{Relevant and potentially informative} Who is Miss Universe Guyana 2017?

    \end{itemize}
\end{itemize}
\begin{itemize}
    \item \textbf{Claim} You will learn more about Donald Trump by going down to the Federal Election Commission to see the financial disclosure form than by looking at tax returns.
    \begin{itemize}
        \item \textbf{Relevant and uninformative} Where is Donald Trump's tax return?
        \item \textbf{Relevant and weakly informative} How can you learn more about Donald Trump by looking at tax returns?
        \item \textbf{Relevant and weakly informative} How can you learn more about Donald Trump by going down to the Federal Election Commission?
        \item \textbf{Relevant and potentially informative} How much does Donald Trump donate to charity?
        \item \textbf{Relevant and informative} What is Donald Trump's tax rate?
    \item \textbf{Relevant and informative} What type of taxes does Donald Trump pay?

    \end{itemize}
\end{itemize}
\begin{itemize}
\item \textbf{Claim} If Congress fails to act the Obama administration intends to give away control of the internet to an international body akin to the United Nations.
    \begin{itemize}
    \item \textbf{Relevant and uninformative} What constitutes an international body?
    \item \textbf{Relevant and uninformative} What would happen if Congress did not act?
    \item \textbf{Relevant and weakly informative} Who has oversight over the internet in America?
    \item \textbf{Relevant and weakly informative} What countries have officers involved in the Internet Corporation for Assigned Names and Numbers?
    \item \textbf{Relevant and weakly informative} What is the United Nations?
    \item \textbf{Relevant and potentially informative} What did the Obama administration intend to do with control of the internet? 
    \item \textbf{Relevant and informative question} What organization does the Obama administration want to give control of the internet to?

    \end{itemize}
\end{itemize}

\subsection{Prerequisites for the different criteria}
\begin{enumerate}
    \item Intelligibility should be judged without looking at the claim.
    \item A clear question should be intelligible. When assessing the clarity, the claim can be checked for more details.
    \item Relevance and informativeness should be annotated by looking at the claim. 
    \item A relevant question needs to be intelligible.
   \item An informative question is intelligible, clear, and relevant.
   
\end{enumerate}





    

\section{Implementation Details}
\subsection{Pre-processing} We remove duplicates by setting a 0.8 BLEU threshold so that extremely similar questions are removed before the re-ranking.


\subsection{Models and Hyperparameters} 

We use the standard HuggingFace implementation of BART (\url{https://huggingface.co/transformers/model_doc/bart.html}) and BERT (\url{https://huggingface.co/transformers/model_doc/bert.html}).

\subsubsection{Generator}
\textbf{BartForConditionalGeneration
}

\textbf{Encoder layers} 12

\textbf{Encoder heads} 16

\textbf{Decoder layers} 12

\textbf{Decoder layers} 16

\textbf{Dimensionality} 1024

\textbf{Feed-forward layers dimensionality} 4096

\textbf{Activation function} gelu

\textbf{Dropout} 0.1

\textbf{Batch size} 2 per device

\textbf{Early stopping patience} 10

\subsubsection{Re-ranker}

\textbf{BertForSequenceClassification
}

\textbf{Number of labels} 1 (regression)

\textbf{Encoder layers} 12

\textbf{Encoder heads} 12

\textbf{Dimensionality} 768

\textbf{Feed-forward layers dimensionality} 3072

\textbf{Dropout} 0.1

\textbf{Activation function} gelu

\textbf{Batch size} 128 per device

\textbf{Early stopping patience} 10

\end{document}